\documentclass{article}
\usepackage[preprint]{colm2026_conference}

\usepackage{microtype}
\usepackage{hyperref}
\usepackage{url}
\usepackage{booktabs}
\usepackage{graphicx}
\usepackage{amsmath}
\usepackage{amssymb}
\usepackage{multirow}
\usepackage[table]{xcolor}
\usepackage{subcaption}
\usepackage{float}

\usepackage{amsmath,amsfonts,bm}

\def\eqref#1{equation~\ref{#1}}

\def\1{\bm{1}}

\DeclareMathAlphabet{\mathsfit}{\encodingdefault}{\sfdefault}{m}{sl}
\SetMathAlphabet{\mathsfit}{bold}{\encodingdefault}{\sfdefault}{bx}{n}

\definecolor{darkblue}{rgb}{0, 0, 0.5}
\hypersetup{colorlinks=true, citecolor=darkblue, linkcolor=darkblue, urlcolor=darkblue}

\title{Riding Brainwaves in LLM Space: \\
Understanding Activation Patterns Using Individual Neural Signatures}

\author{%
  Ajan Subramanian\thanks{Equal contribution.} \quad
  Sumukh Bettadapura\footnotemark[1] \quad
  Rohan Sathish \\
  Kubo Technologies
}

\begin{document}

\lhead{}

\maketitle

\begin{abstract}
Consumer-grade EEG is entering everyday devices, from earbuds to headbands, raising the question of whether language models can be adapted to individual neural signatures.
We test this by asking whether frozen LLM representations encode person-specific EEG signals, directions in activation space that predict one person's brain activity but not another's.
Using word-level EEG from 30 participants reading naturalistic sentences (ZuCo corpus), we train a separate linear probe for each person, mapping hidden states from a frozen Qwen 2.5 7B to that individual's EEG power.
Person-specific probes outperform a single population probe on every EEG feature tested; for high-gamma power, the person-specific probe achieves $\rho = 0.183$, a ninefold improvement over the population probe ($\rho = 0.020$, $p < 10^{-4}$).
A negative control, fixation count, shows no person-specific advantage ($p = 0.360$); fixation count reflects word length and frequency rather than individual cognition.
The individual directions are temporally stable (split-half cosine $= 0.824$), non-transferable across people (self $\rho = 0.369$ vs.\ other $\rho = 0.143$, $p < 10^{-19}$), and distinct from the shared population signal: person-specific probes retain predictive power after the population component is removed.
The person-specific signal concentrates in the model's deep layers, rising consistently with depth and peaking at Layer~24 of 28.
The results are consistent across architectures (LLaMA 3.1 8B) and survive word-level confound controls.
Frozen language models contain stable, person-specific neural directions in their deep layers, providing a geometric foundation for EEG-driven personalization.
\end{abstract}

\section{Introduction}
\label{sec:intro}

When two people read the same word, their brains respond differently.
Precision neuroimaging has shown that this individual variation is not noise: it is a dominant source of variance in how brains are organized, exceeding task and session effects~\citep{gratton2018functional, visconti2025individual}.
As EEG hardware moves from the laboratory to consumer devices, individual neural variation is becoming not just a scientific question but a practical one: can systems adapt to the person wearing the headset?
We ask whether these individual differences have structure in the geometry of large language models.

The standard approach to comparing language models with brains averages neural data across participants and evaluates at the population level~\citep{schrimpf2021neural, caucheteux2022brains, goldstein2022shared, toneva2019interpreting}.
This has been productive, but it retains only what is shared across people and discards what is specific to any one of them.
A separate line of work has shown that linear directions in LLM activation space encode high-level concepts such as truthfulness and sentiment~\citep{zou2023representation, turner2023steering}, but these directions are defined by what the text says, not by who reads it.
Neither tradition has asked whether language models encode properties of the individual processing the text.

We test this directly.
The key idea is simple: instead of training a single linear probe that maps LLM hidden states to the average neural signature across all participants, we train a separate probe for each person.
If the model's representations contain individual-specific structure, each person's probe should find a different direction in activation space.
That direction should predict that person's neural signatures better than a one-size-fits-all mapping.
We apply this to the ZuCo corpus~\citep{hollenstein2018zuco, hollenstein2020zuco}, which provides word-level EEG and eye-tracking from 30 participants reading naturalistic sentences, using PCA-reduced hidden states from a frozen Qwen 2.5 7B as the representation space.

Our contribution is to show that frozen language model representations encode individual-level neural structure: person-specific linear probes find distinct directions in activation space that are temporally stable, non-transferable across people, and separable from the shared population signal.
We validate this across two architectures, two datasets, and multiple negative controls.

The remainder of the paper presents the experimental setup (\S\ref{sec:setup}), core results and characterization (\S\ref{sec:results}), robustness and generality tests (\S\ref{sec:robustness}), and discussion (\S\ref{sec:discussion}).

\section{Related Work}
\label{sec:related}

\paragraph{Representation engineering.}
The Linear Representation Hypothesis, that high-level concepts are encoded as linear directions in activation space, has received growing support~\citep{park2023linear, nanda2023emergent, elhage2022toy}.
Representation engineering extracts such directions and uses them to read or steer model behavior~\citep{zou2023representation}: Activation Addition~\citep{turner2023steering} and its contrastive variant~\citep{panickssery2023steering} steer generation along concept axes, and inference-time intervention elicits truthful outputs~\citep{li2023inference, marks2023geometry}.
In all of this work, the target concept is defined by text: ``truthful'' versus ``false'' statements, ``helpful'' versus ``harmful'' completions.

\paragraph{Neuro-linguistic alignment.}
A parallel literature maps LLM representations onto neural data.
Brain-Score benchmarks~\citep{schrimpf2021neural} and related work~\citep{caucheteux2022brains, goldstein2022shared, toneva2019interpreting, jain2018incorporating} find that mid-layer representations best predict population-averaged neural signals during language comprehension. Scaling improves this alignment~\citep{gao2025scaling}, though the role of instruction tuning is debated~\citep{aw2023instruction, tuckute2024driving}.
This paradigm has been productive, but it evaluates alignment at the population level.

\paragraph{Precision neuroimaging.}
Recent precision neuroimaging has established that individual-level variation in brain function is substantial.
Functional connectivity fingerprints are unique and stable across sessions, predicting individual cognitive abilities~\citep{finn2015fingerprinting}.
Individual-level factors exceed task-level and daily variance in functional brain networks~\citep{gratton2018functional}, and individual differences shape conceptual representation in brain regions where personal knowledge meets sensory input~\citep{visconti2025individual}.

\paragraph{Word-level EEG prediction.}
At the intersection of these two fields, a growing body of work predicts word-level neural signals from LLM representations: information-theoretic measures predict ERP amplitudes~\citep{frank2015erp}, LLM surprisal explains N400 effects~\citep{michaelov2024strong}, and cognitive evaluation frameworks benchmark embeddings against EEG~\citep{hollenstein2019cognival, hollenstein2023zuco}.

\medskip\noindent
Across these four traditions, neural prediction targets population-averaged signals or text-defined concepts.
Whether language model representations also encode structure specific to individual readers has not been explored.

\section{Experimental Setup}
\label{sec:setup}

\begin{figure*}[!ht]
\centering
\includegraphics[width=0.92\textwidth]{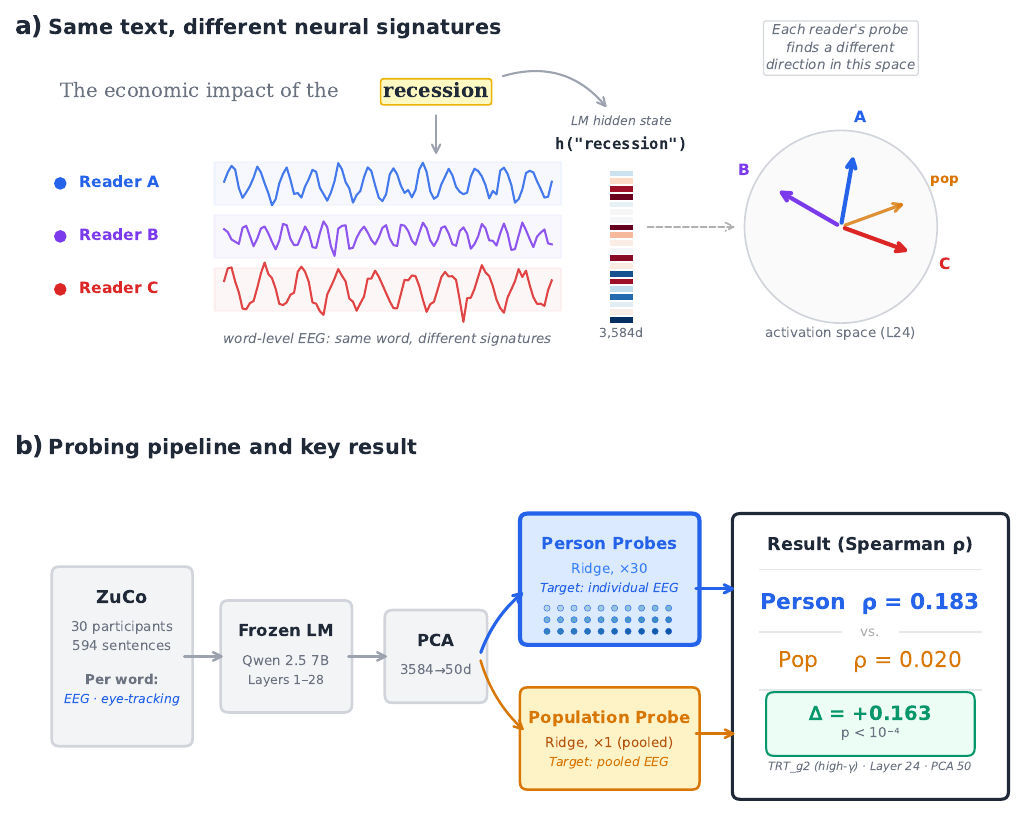}
\caption{\textbf{(a)}~Different readers produce distinct neural signatures for the same word; each person-specific probe learns a different direction in the LM's activation space (``pop'' = population). \textbf{(b)}~Probing pipeline: words pass through a frozen LM and PCA; 30 person-specific Ridge probes are compared against a single population probe. The person-specific probe achieves $\bar{\rho} = 0.183$ vs.\ $0.020$ for the population probe on TRT\_g2 (high-$\gamma$, Layer~24).}
\label{fig:pipeline}
\end{figure*}

If frozen language models encode individual-level neural structure, a simple test should reveal it: train one linear probe per person and check whether it outperforms a single probe trained on everyone pooled.
We implement this as follows (Figure~\ref{fig:pipeline}).

\subsection{Data}
\label{sec:data}

We combine ZuCo 1.0~\citep{hollenstein2018zuco} (12 participants, 300 sentences) and ZuCo 2.0~\citep{hollenstein2020zuco} (18 participants, 349 sentences), yielding 30 participants, 594 unique sentences, and 196{,}162 word-level observations.
Both corpora provide simultaneous EEG and eye-tracking during naturalistic sentence reading.
For each fixated word, EEG power is decomposed into eight frequency sub-bands (theta through high-gamma) across five temporal reading windows: First Fixation Duration (FFD), Gaze Duration (GD), Total Reading Time (TRT), Single Fixation Duration (SFD), and Go-Past Time (GPT), yielding $5 \times 8 = 40$ EEG features per word per participant.
All values are averaged across 105 electrode channels.
We also extract mean pupil size and fixation count (nFixations) per word.
Mean pupil size varies across individuals due to physiological factors such as age and baseline arousal, making it a natural person-specific target.
Fixation count depends primarily on word length and frequency rather than on who is reading, so it serves as a negative control: if person-specific probes show no advantage for fixation count, the advantage for EEG is not simply per-participant noise.
Full dataset details are in Appendix~\ref{app:data-details}.

\subsection{Hidden-State Extraction}
\label{sec:extraction}

We extract hidden states from Qwen 2.5 7B Instruct~\citep{qwen2025qwen25technicalreport}, a 28-layer decoder with 3{,}584-dimensional representations.
Each sentence is passed through the frozen model; we extract hidden states at layers $\ell \in \{1, 4, 8, 12, 16, 20, 24, 28\}$.
For multi-token words, we average constituent token representations.
The raw hidden dimension (3{,}584) is too large for reliable regression with the available data (${\sim}5{,}000$ words per participant), so we reduce it via Principal Component Analysis (PCA), which selects the directions of greatest variance.
We fit PCA on all 12{,}200 unique words and project to $d = 50$ dimensions per layer.
For replication, we run the same pipeline on LLaMA 3.1 8B Instruct~\citep{grattafiori2024llama} (32 layers, 4{,}096-dim) at layers $\{1, 4, 8, 12, 16, 20, 24, 28, 32\}$.

\subsection{Probing Framework}
\label{sec:probing}

A probe is a linear regression that maps the model's hidden-state representation of a word to the neural signature that word elicited.
We say a person-specific probe \emph{outperforms} the population probe if it achieves a higher Spearman correlation on held-out data for that participant.

For each participant $i$ and EEG feature $f$, we train a person-specific Ridge regression:
\begin{equation}
\hat{y}_{i,f}(w) = \mathbf{x}(w)^\top \boldsymbol{\beta}_{i,f} + b_{i,f}
\label{eq:person-probe}
\end{equation}
where $\mathbf{x}(w) \in \mathbb{R}^{50}$ is the PCA-reduced hidden state of word $w$, $\boldsymbol{\beta}_{i,f} \in \mathbb{R}^{50}$ is the learned weight vector, $b_{i,f}$ is a scalar bias, and $\hat{y}_{i,f}(w)$ is the predicted EEG power for participant $i$ on feature $f$.
The Ridge regularization strength $\alpha \in \{0.01, 0.1, 1, 10, 100, 1000\}$ is selected by held-out validation.
As a baseline, a single population probe is trained on all 30 participants pooled:
\begin{equation}
\hat{y}_{\text{pop},f}(w) = \mathbf{x}(w)^\top \boldsymbol{\beta}_{\text{pop},f} + b_{\text{pop},f}
\label{eq:pop-probe}
\end{equation}
We evaluate via 5-fold cross-validation, splitting at the sentence level (all words from a given sentence appear in the same fold) to prevent within-sentence context leakage.
We report Spearman $\rho$ between predicted and actual values at the word level, averaged across folds and participants.
Person-specific $\bar{\rho}$ versus population $\bar{\rho}$ is tested via paired $t$-test across the 30 participants.

\subsection{Confound Features}
\label{sec:confound-features}

Each word is annotated with log word frequency, word length, normalized sentence position, and GPT-2 surprisal.
These capture the major lexical confounds known to modulate both EEG and eye-tracking signals and are used in \S\ref{sec:confounds} to test whether the person-specific advantage survives after these word-level properties are regressed out.

\section{Results}
\label{sec:results}

\subsection{Person-Specific Probes Outperform Population Probes}
\label{sec:results-core}

Table~\ref{tab:main-results} reports person-specific and population probe performance across all 40 EEG features at the best configuration (Layer~24, PCA~50).

\begin{table}[!ht]
\centering
\small
\begin{tabular}{lcccc}
\toprule
\textbf{Feature} & \textbf{Person $\rho$} & \textbf{Pop $\rho$} & \textbf{$\Delta$} & \textbf{$p$} \\
\midrule
meanPupilSize        & 0.261 & 0.071 & +0.189 & $< 10^{-4}$ \\
TRT\_g2 (gamma high) & 0.183 & 0.020 & +0.163 & $< 10^{-4}$ \\
TRT\_g1 (gamma low)  & 0.156 & 0.043 & +0.113 & $< 10^{-4}$ \\
TRT\_b2 (beta high)  & 0.145 & 0.069 & +0.077 & $< 10^{-4}$ \\
TRT\_t1 (theta low)  & 0.100 & 0.024 & +0.075 & $< 10^{-4}$ \\
TRT\_a1 (alpha low)  & 0.127 & 0.054 & +0.073 & $< 10^{-4}$ \\
FFD\_t1              & 0.090 & 0.023 & +0.067 & $< 10^{-4}$ \\
FFD\_a1              & 0.117 & 0.053 & +0.064 & $< 10^{-4}$ \\
\midrule
TRT (reading time)   & 0.269 & 0.247 & +0.023 & $3.0 \times 10^{-5}$ \\
nFixations           & 0.234 & 0.231 & +0.004 & 0.360 \\
\bottomrule
\end{tabular}
\caption{Person-specific vs.\ population probe performance (Spearman $\rho$) at Layer~24, PCA~50. All EEG comparisons are significant at $p < 10^{-4}$ (paired $t$-test, $N = 30$). nFixations (a text-structural measure) shows no person-specific advantage ($p = 0.360$).}
\label{tab:main-results}
\end{table}

Person-specific probes outperform the population probe on every feature, with all comparisons significant at $p < 10^{-4}$ ($N = 30$).
The largest person-specific advantage is for mean pupil size ($\Delta = +0.189$), consistent with pupil diameter being shaped by physiological factors such as age and baseline arousal that vary across individuals.
Among EEG features, the advantage is largest in gamma ($\Delta = 0.113$--$0.163$) and beta ($\Delta = 0.077$), and smallest in theta.
Gamma and beta oscillations are associated with active semantic processing and cortical binding~\citep{buzsaki2004oscillations, engel2001dynamic}, suggesting the person-specific signal reflects higher-order cognition rather than low-level perception.
In word-level EEG prediction, where population-level correlations typically fall in the $0.02$--$0.07$ range (Table~\ref{tab:main-results}), person-specific $\rho$ values of $0.10$--$0.18$ represent a substantial individual signal.

\begin{figure}[!ht]
\centering
\includegraphics[width=\linewidth]{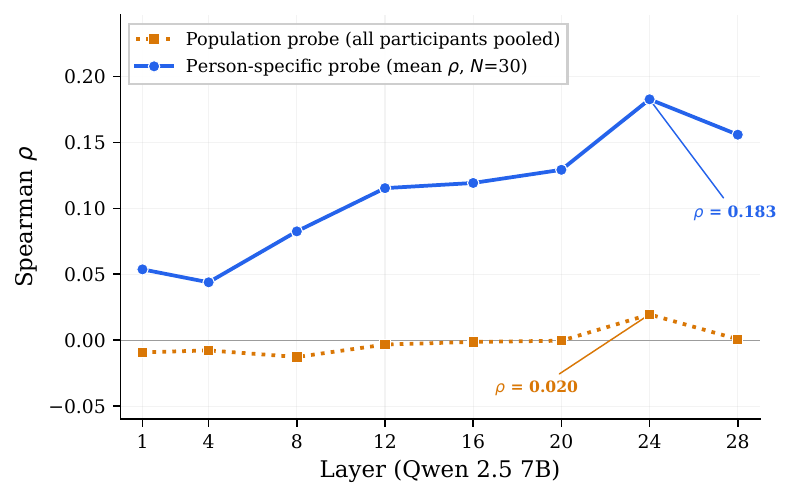}
\caption{Layer-wise probe performance on Qwen 2.5 7B (ZuCo, TRT\_g2). Person-specific probes (solid blue, mean $\rho$ across 30 participants) increase with depth, peaking at Layer~24. The population probe (dashed orange) remains near zero throughout. Layer~28 shows a slight decline relative to Layer~24.}
\label{fig:dissociation}
\end{figure}

\label{sec:dissociation}
Person-specific performance increases consistently with layer depth, peaking at Layer~24 (Figure~\ref{fig:dissociation}).
The population probe remains near zero at every layer, confirming that the individual signal is encoded in directions that a single shared probe cannot recover.
The final layer (28) shows a slight decline, though it still outperforms all layers below 20; this is expected because PCA~50 retains only 69\% of variance at this depth (vs.\ $>$99\% at layers 4--24), so some person-specific signal is lost to dimensionality reduction.

Fixation count (nFixations) serves as a negative control.
The number of times a reader fixates a word depends primarily on word length and frequency, not on who is reading.
If the person-specific advantage were an artifact of per-participant noise fitting, we would expect it to appear for fixation count as well.
It does not: person-specific and population probes perform nearly identically ($\rho = 0.234$ vs.\ $0.231$, $\Delta = 0.004$, $p = 0.360$), confirming the advantage is specific to neural signals.

\subsection{Characterization: Genuine Individual Signatures}
\label{sec:characterization}

Several alternative explanations must be ruled out.
All characterization results use the best configuration (Layer~24, PCA~50, meanPupilSize).

\begin{figure}[!ht]
\centering
\includegraphics[width=0.75\linewidth]{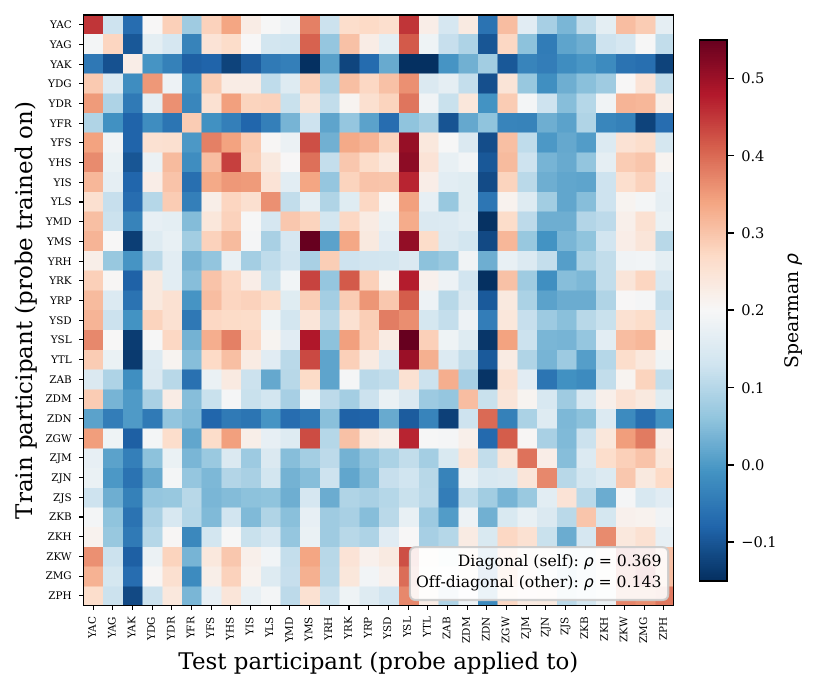}
\caption{Cross-person transfer matrix (Qwen 2.5 7B, mean pupil size, Layer~24, PCA~50). Each cell $(i, j)$ shows Spearman $\rho$ when participant~$i$'s probe predicts participant~$j$'s mean pupil size. The diagonal (self $\rho = 0.369$) is consistently higher than the off-diagonal (other $\rho = 0.143$), confirming person-specificity. LLaMA 3.1 8B shows the same pattern (Table~\ref{tab:llama-results}).}
\label{fig:transfer-matrix}
\end{figure}

\paragraph{Cross-person transfer.}
If person-specific directions are genuinely individual, a probe trained on one participant should fail when applied to another.
We test this with a full $30 \times 30$ transfer matrix: participant~$i$'s probe evaluated on participant~$j$'s data.
Self-prediction averages $\rho = 0.369$, while cross-person prediction drops to $\rho = 0.143$ ($p = 5.8 \times 10^{-20}$, paired $t$-test; Figure~\ref{fig:transfer-matrix}).
Each probe captures information specific to the person it was trained on.
A small number of participant pairs show moderate cross-person correlation (visible as off-diagonal warm cells), suggesting partial overlap in neural processing style between those individuals.
Geometrically, the learned weight vectors confirm this pattern: the mean pairwise cosine similarity across all $\binom{30}{2} = 435$ participant pairs is $0.351$, and cosine to the population probe averages $0.228$: the directions share some common structure but occupy distinct regions of activation space.

\paragraph{Independence from population signal.}
A subtler concern is that each person's probe might learn a slightly different weighting of the same population-level mapping rather than a genuinely different direction.
To test this, we subtract the population probe's prediction from each participant's actual EEG and re-evaluate the person-specific probe on the residuals.
If the person-specific direction were merely a rescaled version of the population direction, it should fail on these residuals.
It does not: the person-specific probes retain significant predictive power on the residuals ($p < 10^{-4}$), confirming that they capture signal distinct from the population mapping.

\paragraph{Temporal stability.}
We split each participant's reading session into temporal halves and train independent probes on each half.
The cosine similarity between the two weight vectors averages $0.824$ ($p = 6.6 \times 10^{-31}$).
Whatever these probes are finding, it persists across the reading session.

\section{Robustness and Generality}
\label{sec:robustness}

We now test whether the person-specific advantage survives confound removal, holds up under targeted negative controls, and replicates in a second model architecture.

\subsection{Word-Level Confound Controls}
\label{sec:confounds}

Word-level EEG power correlates with basic lexical properties: frequent words are read faster, longer words attract more fixations, and word predictability (surprisal) modulates processing effort.
If person-specific probes merely learned that ``participant A reads frequent words more quickly than participant B,'' the advantage would be an artifact of per-person reading speed interacting with word difficulty.

To test this, we regress out four word-level confounds (log word frequency, word length, sentence position, and GPT-2 surprisal) from each EEG feature on a per-participant basis, then retrain the full probe pipeline on the residual signal, the part of each person's EEG that cannot be predicted from word properties alone.
The nuisance regressions explain very little variance (mean $R^{2} < 0.01$ for all EEG bands; $R^{2} = 0.045$ for meanPupilSize), confirming that word properties are only weakly related to the EEG signals the probes predict.

\begin{table}[!ht]
\centering
\small
\begin{tabular}{lcccc}
\toprule
\textbf{Feature} & \textbf{Person $\rho$} & \textbf{Person $\rho$} & \textbf{Pop $\rho$} & \textbf{$p$} \\
& \textbf{(raw)} & \textbf{(confounds removed)} & \textbf{(confounds removed)} & \textbf{(confounds removed)} \\
\midrule
TRT\_g2       & 0.183 & 0.169 & 0.057 & $8.2 \times 10^{-8}$ \\
TRT\_g1       & 0.151 & 0.141 & 0.065 & $3.5 \times 10^{-6}$ \\
meanPupilSize & 0.258 & 0.181 & 0.118 & $1.9 \times 10^{-7}$ \\
TRT\_b2       & 0.153 & 0.124 & 0.068 & $3.1 \times 10^{-5}$ \\
TRT\_a1       & 0.125 & 0.107 & 0.070 & $3.5 \times 10^{-5}$ \\
\bottomrule
\end{tabular}
\caption{Person-specific probe performance before and after removing word-level confounds (word frequency, length, sentence position, GPT-2 surprisal) via per-participant nuisance regression. The person-specific advantage survives for all five features ($p < 10^{-4}$). Raw values come from re-running probes within the confound-control pipeline (same CV folds as the residualized condition) and may differ slightly from Table~\ref{tab:main-results}.}
\label{tab:confounds}
\end{table}

The person-specific advantage survives confound removal for every feature ($p < 10^{-4}$; Table~\ref{tab:confounds}), confirming the probes detect individual neural signatures that persist after word-level properties have been regressed out.

\subsection{Specificity Controls}
\label{sec:specificity}

A second class of concern is statistical: perhaps \emph{any} per-person noisy signal would produce a similar advantage, simply because individual probes overfit to per-participant noise.
We address this with six targeted controls (Table~\ref{tab:controls}); the cross-person transfer test (\S\ref{sec:characterization}) serves as a seventh.

\begin{table}[!ht]
\centering
\small
\begin{tabular}{lcccc}
\toprule
\textbf{Condition} & \textbf{Person $\rho$} & \textbf{Pop $\rho$} & \textbf{$\Delta$} & \textbf{$p$} \\
\midrule
Full model (TRT\_g2)     & 0.183 & 0.020 & +0.163 & $3.1 \times 10^{-9}$ \\
Shuffled (word$\leftrightarrow$EEG scrambled) & 0.001 & 0.001 & ${\approx}$0.000 & $^{\dagger}$ \\
nFixations (neg.\ ctrl.) & 0.234 & 0.231 & +0.004 & 0.360 \\
Random proj.\ (50 random dirs.) & 0.179 & 0.014 & +0.165 & $^{\ddagger}$ \\
GloVe 50d (static, no LLM)      & 0.057 & $-$0.006 & +0.063 & $4.8 \times 10^{-7}$ \\
Random emb.\ (no LLM)          & 0.010 & 0.004 & +0.006 & $^{\S}$ \\
\bottomrule
\end{tabular}
\caption{Specificity controls for TRT\_g2 at Layer~24, PCA~50. \emph{Shuffled}: word-to-EEG pairings randomly scrambled (10 permutations). \emph{Random proj.}: PCA's top-50 directions replaced with 50 random orthogonal directions. \emph{GloVe 50d}: LLM activations replaced with static GloVe word vectors~\citep{pennington2014glove}. \emph{Random emb.}: LLM activations replaced with fixed random vectors per word. $^{\dagger}$Both collapse to zero; no meaningful comparison. $^{\ddagger}$Single run; paired test not applicable. $^{\S}$Both collapse to near zero; no meaningful comparison.}
\label{tab:controls}
\end{table}

The \emph{shuffled} control scrambles which neural signature is paired with which word (10 permutations), and performance collapses to zero ($\rho \approx 0.001$), confirming the probes require correct word-to-response alignment.
The \emph{nFixations} control uses a text-structural feature (number of fixations) that shows no person-specific advantage ($\Delta = 0.004$, $p = 0.360$), confirming the effect is specific to neural signals.
The \emph{random-projection} baseline replaces PCA's top-50 directions with 50 random orthogonal directions; performance is nearly identical ($\Delta = 0.165$ vs.\ $0.163$), indicating the person-specific signal is distributed broadly across the activation space.

The final two controls test whether the advantage requires LLM representations at all.
The \emph{random-embedding} control replaces LLM activations with fixed random vectors per word ($\mathcal{N}(0, 1)$ in $\mathbb{R}^{50}$): both probes collapse to near zero ($\rho \approx 0.01$, $\Delta = 0.006$, $p = 0.203$), ruling out regression artifacts.
The \emph{GloVe} control~\citep{pennington2014glove} uses static 50-dimensional word embeddings: a small but significant advantage emerges ($\Delta = 0.063$, $p = 4.8 \times 10^{-7}$), yet it is $2.6\times$ smaller than the LLM's ($\Delta = 0.163$) and absolute accuracy is $3.2\times$ lower ($\rho = 0.057$ vs.\ $0.183$).
Together, these controls establish a clear hierarchy: random vectors (no effect) $<$ static embeddings (small effect) $\ll$ contextual LLM representations (large effect).

\subsection{Second Model: LLaMA 3.1 8B}
\label{sec:llama}

If person-specific directions are a property of how decoder transformers represent text in general, the finding should not depend on Qwen.
We replicate the full pipeline on LLaMA 3.1 8B Instruct~\citep{grattafiori2024llama} (32 layers, 4{,}096-dim hidden states), extracting activations at layers 1, 4, 8, 12, 16, 20, 24, 28, and 32.

\begin{table}[!ht]
\centering
\small
\begin{tabular}{lcc}
\toprule
\textbf{Metric} & \textbf{Qwen 2.5 7B} & \textbf{LLaMA 3.1 8B} \\
\midrule
Best layer                & 24            & 24 \\
TRT\_g2 person $\rho$    & 0.183         & 0.196 \\
TRT\_g2 pop $\rho$       & 0.020         & 0.012 \\
meanPupilSize person $\rho$ & 0.261      & 0.278 \\
Self transfer $\rho$     & 0.369         & 0.384 \\
Other transfer $\rho$    & 0.143         & 0.154 \\
Split-half cosine        & 0.824         & 0.841 \\
\bottomrule
\end{tabular}
\caption{Side-by-side comparison of person-specific probe metrics for Qwen 2.5 7B and LLaMA 3.1 8B (both at Layer~24, PCA~50). All key metrics replicate quantitatively.}
\label{tab:llama-results}
\end{table}

The replication is quantitatively close (Table~\ref{tab:llama-results}): LLaMA peaks at the same depth (Layer~24), produces a nearly identical person-specific advantage ($\Delta = 0.183$ for both models), and matches Qwen on cross-person transfer (self $\rho = 0.384$ vs.\ other $\rho = 0.154$, $p < 10^{-19}$) and split-half reliability (cosine $= 0.841$).
The person-specific directions emerge at the same depth, with the same magnitude, in two independently trained architectures.

\subsection{Cross-Dataset Validation}
\label{sec:cross-dataset}

Our primary results pool ZuCo 1.0 (12 participants) and ZuCo 2.0 (18 participants), collected in separate sessions with different stimuli.
To test generalization, we train population probes on one corpus and evaluate on the other.
Transfer varies by feature type: nFixations (text-structural) retains 97\% (cross $\rho = 0.226$ vs.\ within $\rho = 0.233$); mean pupil size retains 61\%; and high-frequency EEG bands retain 26--35\%, consistent with these features being the most person-specific.

\section{Discussion}
\label{sec:discussion}

\paragraph{A mechanistic finding.}
This work establishes that frozen language model representations encode individual-level neural structure accessible through simple linear probes.
The finding is mechanistic: person-specific directions exist in the deep layers of the network, are temporally stable, non-transferable across people, and capture signal that persists after the population component is removed.
A practical consequence for brain-alignment benchmarks such as Brain-Score~\citep{schrimpf2021neural} follows directly: evaluating at the population level retains only the coarsest shared signal, while person-specific probing recovers substantially more of the decodable information.

\paragraph{The individual signal is encoded in directions, not layers.}
The layer profile (Figure~\ref{fig:dissociation}) clarifies the structure of this encoding.
The population probe remains near zero at every layer, not just at the wrong depth.
The individual signal is encoded in directions that pooling across participants destroys.
This distinction matters: the barrier to capturing individual variation is not about choosing the right layer, but about training separate mappings for each person.

\paragraph{Implications for personalization.}
Person-specific linear directions in activation space open the possibility of biologically grounded personalization.
Current personalization methods rely on text-based proxies such as user profiles or preference histories~\citep{zhang2024personalization}; person-specific neural directions offer a complementary signal that is grounded in how the individual's brain processes language.
As EEG hardware moves from the laboratory to consumer devices, brief calibration sessions could in principle extract a person's neural direction in model space, creating a compact individual signature.
Inference-time steering methods~\citep{turner2023steering, zhao2025steerx} could then use this direction to adapt generation to the person wearing the headset.

The present work identifies the directions but does not inject them.
The mechanistic next step is causal intervention: adding person-specific directions to the residual stream during generation and measuring whether this produces meaningful behavioral changes.
Whether these directions can be extracted from brief calibration sessions and mapped onto a shared representational space is a natural follow-up question.

\paragraph{Limitations.}
Several limitations should temper interpretation.
Our probes are linear, and nonlinear methods might capture additional structure, though they risk overfitting at the current data scale (roughly 5{,}000 words per participant).
We do not attempt causal interventions: the person-specific directions are identified, not injected into the residual stream.
The ZuCo dataset provides 30 participants, which is standard for simultaneous EEG and eye-tracking but modest for claims about the generality of individual differences.
EEG power is computed during fixation windows (FFD, GD, TRT), so the EEG signal is not fully decoupled from fixation mechanics; the nFixations control addresses this partially, but a more thorough disentanglement would require time-locked analysis independent of fixation structure.
Finally, all EEG features are averaged across 105 electrodes; electrode-level analysis could reveal whether the effect is concentrated in language-related brain regions.
Two natural extensions are causal intervention (injecting person-specific directions during generation to adapt model behavior) and electrode-level analysis linking the computational finding to known language-processing areas.

\section{Conclusion}
\label{sec:conclusion}

Frozen decoder language models contain person-specific linear directions that predict individual neural signatures during naturalistic reading.
These directions are temporally stable, non-transferable across people, and distinct from the shared population signal.
They concentrate in the deepest layers of the network and emerge consistently across two independently trained architectures.
As EEG hardware becomes portable, this geometry offers a path toward neural personalization: locating an individual's direction in model space from a brief calibration session and steering generation accordingly.

\appendix
\section{LLM Usage Declaration}
\label{app:llm-declaration}

This work uses two pretrained language models as objects of study: Qwen 2.5 7B Instruct~\citep{qwen2025qwen25technicalreport} and LLaMA 3.1 8B Instruct~\citep{grattafiori2024llama}.
No models were fine-tuned; all experiments extract hidden states from frozen forward passes.
LLM-based code assistants were used for experiment scripting and data analysis pipelines.
LLM agents were used to draft and refine the manuscript text, search for and format references, and generate figure layouts.
All results were verified by the authors against raw experimental outputs.
The scientific claims, experimental design, and interpretations were authored by the research team.

\section{Ethics Statement}
\label{app:ethics}

Both datasets are publicly available under institutional ethical approval obtained by the original collectors.
The ZuCo 1.0 corpus~\citep{hollenstein2018zuco} was collected at the University of Zurich with ethics board approval; ZuCo 2.0~\citep{hollenstein2020zuco} was collected under the same framework.
All participants provided informed consent.
Our work involves no new human data collection.

Cognitive-load detection technology could potentially be misused for attention surveillance or manipulative content optimization.
The person-specific neural signatures we identify are derived from controlled laboratory recordings and cannot currently be extracted from naturalistic interaction.
We encourage responsible deployment of bioadaptive systems with informed consent and transparent user controls.

\section{Reproducibility Details}
\label{app:reproducibility}

\paragraph{Model identifiers.}
Qwen 2.5 7B Instruct: \texttt{Qwen/Qwen2.5-7B-Instruct}.
LLaMA 3.1 8B Instruct: \texttt{meta-llama/Llama-3.1-8B-Instruct}.
All accessed via HuggingFace Transformers~\citep{wolf2020transformers}.

\paragraph{PCA and probe details.}
PCA is fit on the full set of 12{,}200 unique words per layer, projecting to 50 dimensions.
Ridge regression $\alpha$ is selected from $\{0.01, 0.1, 1, 10, 100, 1000\}$ via held-out validation on fold~1.
5-fold cross-validation is sentence-stratified: all words from a given sentence appear in the same fold.

\paragraph{Hardware.}
Embedding extraction and probe training were performed on an NVIDIA L4 GPU (24~GB VRAM) with CUDA 12.8.
Ridge regression was accelerated via \texttt{cuml.accel} dispatching to GPU.
Total compute for the full probe sweep (8 layers $\times$ 2 PCA dimensions $\times$ 30 participants $\times$ 40+ features) was approximately 11--17 hours with 6-worker parallelization.

\paragraph{Software.}
Python 3.11, PyTorch 2.7.1, Transformers, NumPy, Pandas, SciPy, scikit-learn, cuML 25.12.

\section{Data Details}
\label{app:data-details}

\paragraph{ZuCo 1.0.}
The Zurich Cognitive Language Processing Corpus~\citep{hollenstein2018zuco} recorded 12 native English-speaking participants reading 300 Wikipedia and movie-review sentences under a natural reading task (task2-NR).
Participants were healthy adults with normal or corrected-to-normal vision.
EEG was recorded at 500~Hz from 105 electrodes using a BioSemi ActiveTwo system; eye movements were recorded simultaneously with an EyeLink 1000 Plus at 1000~Hz.

\paragraph{ZuCo 2.0.}
A second recording~\citep{hollenstein2020zuco} collected data from 18 additional participants reading 349 Wikipedia sentences under the same protocol (task1-NR).
Of these, 250 sentences pass quality filtering.
The same hardware and preprocessing pipeline were used.
No overlap exists in participants or stimulus sentences.

\paragraph{EEG frequency bands.}
Eight sub-bands are extracted: theta~1 (t1, 4--6~Hz), theta~2 (t2, 6.5--8~Hz), alpha~1 (a1, 8.5--10~Hz), alpha~2 (a2, 10.5--13~Hz), beta~1 (b1, 13.5--18~Hz), beta~2 (b2, 18.5--30~Hz), gamma~1 (g1, 30.5--40~Hz), and gamma~2 (g2, 40--49.5~Hz).

\paragraph{Reading windows.}
Five temporal windows from eye-tracking are used: First Fixation Duration (FFD), Gaze Duration (GD), Total Reading Time (TRT), Single Fixation Duration (SFD), and Go-Past Time (GPT).

\paragraph{Coverage.}
Words that a participant did not fixate receive no EEG data and are excluded.
EEG coverage ranges from 44\% to 83\% across participants.

\section{Scaling Comparison: 12 vs.\ 30 Participants}
\label{app:scaling}

We originally ran the pipeline on ZuCo 1.0 alone (12 participants).
Scaling to the combined 30-participant dataset strengthened every characterization metric (Table~\ref{tab:scaling}).

\begin{table}[h]
\centering
\small
\begin{tabular}{lccc}
\toprule
\textbf{Metric} & \textbf{$N{=}12$} & \textbf{$N{=}30$} & \textbf{Change} \\
\midrule
TRT\_g2 person $\rho$ & 0.158 & 0.183 & +16\% \\
meanPupilSize person $\rho$ & 0.265 & 0.261 & stable \\
Self transfer $\rho$ & 0.227 & 0.369 & +63\% \\
Other transfer $\rho$ & 0.011 & 0.143 & increased \\
Transfer $p$-value & $3.2{\times}10^{-7}$ & $5.8{\times}10^{-20}$ & $10^{13}{\times}$ stronger \\
Split-half cosine & 0.752 & 0.824 & +10\% \\
\bottomrule
\end{tabular}
\caption{Comparison of characterization metrics at $N{=}12$ (ZuCo 1.0 only) vs.\ $N{=}30$ (combined). All metrics improve or remain stable with more participants.}
\label{tab:scaling}
\end{table}

\section{EEG Band Specificity Analysis}
\label{app:eeg-bands}

\begin{figure}[h]
\centering
\includegraphics[width=0.85\linewidth]{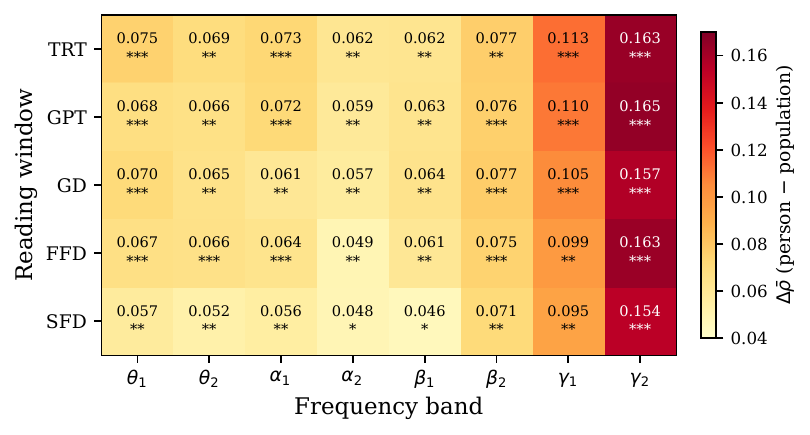}
\caption{Person-specific advantage ($\Delta\bar{\rho} = \bar{\rho}_{\text{person}} - \bar{\rho}_{\text{pop}}$) for 40 EEG features at Layer~24. Stars indicate significance: {*}\,$p<0.001$, {**}\,$p<10^{-4}$, {***}\,$p<10^{-6}$.  Gamma bands carry the strongest individual signal regardless of reading window.}
\label{fig:band-heatmap}
\end{figure}

Figure~\ref{fig:band-heatmap} shows the person-specific advantage ($\Delta\bar{\rho}$) across all 40 EEG features ($5$ reading windows $\times$ $8$ frequency sub-bands).
All 40 cells are significant ($p < 0.001$, paired $t$-test).
Gamma bands ($\gamma_1$, $\gamma_2$) show the strongest person-specific advantage across every reading window, with $\gamma_2$ reaching $\Delta\bar{\rho} > 0.15$ in all five windows.
Lower-frequency bands ($\theta$, $\alpha$) show smaller but still reliable effects ($\Delta\bar{\rho} \approx 0.05$--$0.07$).
Across rows, TRT and GPT windows carry slightly more signal than FFD and SFD.
This pattern connects the computational finding to the neuroscience literature on individual differences in oscillation frequency~\citep{buzsaki2004oscillations, engel2001dynamic, grandy2013individual}: high-gamma activity, which reflects local cortical processing, varies most across individuals.

\section{Effect Sizes and Confidence Intervals}
\label{app:effect-sizes}

\begin{table}[H]
\centering
\small
\begin{tabular}{lcccc}
\toprule
\textbf{Feature} & \textbf{Cohen's $d$} & \textbf{Person $\rho$ 95\% CI} & \textbf{Pop $\rho$ 95\% CI} & \textbf{$p$} \\
\midrule
TRT\_g2        & 1.53 & $[0.155,\; 0.209]$ & $[-0.002,\; 0.040]$ & $3.1{\times}10^{-9}$  \\
meanPupilSize  & 1.99 & $[0.225,\; 0.300]$ & $[0.050,\; 0.093]$  & $9.3{\times}10^{-12}$ \\
TRT\_g1        & 1.17 & $[0.127,\; 0.187]$ & $[0.020,\; 0.066]$  & $5.1{\times}10^{-7}$  \\
TRT\_a1        & 1.18 & $[0.103,\; 0.152]$ & $[0.034,\; 0.075]$  & $4.8{\times}10^{-7}$  \\
TRT\_b2        & 1.08 & $[0.118,\; 0.174]$ & $[0.050,\; 0.087]$  & $2.0{\times}10^{-6}$  \\
\bottomrule
\end{tabular}
\caption{Effect sizes and 95\% bootstrap confidence intervals ($B = 10{,}000$) for the person-specific advantage at Layer~24, Qwen 2.5 7B. Cohen's $d$ is computed from the paired person-vs-population $\rho$ difference across 30 participants. All five features show large effects ($d > 1$); the population probe CI for TRT\_g2 includes zero, confirming that the population signal is negligible for this feature.}
\label{tab:effect-sizes}
\end{table}


\begin{thebibliography}{34}
\providecommand{\natexlab}[1]{#1}
\providecommand{\url}[1]{\texttt{#1}}
\expandafter\ifx\csname urlstyle\endcsname\relax
  \providecommand{\doi}[1]{doi: #1}\else
  \providecommand{\doi}{doi: \begingroup \urlstyle{rm}\Url}\fi

\bibitem[Aw et~al.(2023)Aw, Montariol, AlKhamissi, Schrimpf, and
  Bosselut]{aw2023instruction}
Khai~Loong Aw, Syrielle Montariol, Badr AlKhamissi, Martin Schrimpf, and
  Antoine Bosselut.
\newblock Instruction-tuning aligns {LLM}s to the human brain.
\newblock \emph{arXiv preprint arXiv:2312.00575}, 2023.

\bibitem[Buzs{\'a}ki \& Draguhn(2004)Buzs{\'a}ki and
  Draguhn]{buzsaki2004oscillations}
Gy{\"o}rgy Buzs{\'a}ki and Andreas Draguhn.
\newblock Neuronal oscillations in cortical networks.
\newblock \emph{Science}, 304\penalty0 (5679):\penalty0 1926--1929, 2004.

\bibitem[Caucheteux \& King(2022)Caucheteux and King]{caucheteux2022brains}
Charlotte Caucheteux and Jean-R{\'e}mi King.
\newblock Brains and algorithms partially converge in natural language
  processing.
\newblock \emph{Communications Biology}, 5\penalty0 (1):\penalty0 134, 2022.

\bibitem[Elhage et~al.(2022)Elhage, Hume, Olsson, Schiefer, Henighan, Kravec,
  Hatfield-Dodds, Lasenby, Drain, Chen, Gross, McCandlish, Kaplan, Amodei,
  Wattenberg, and Olah]{elhage2022toy}
Nelson Elhage, Tristan Hume, Catherine Olsson, Nicholas Schiefer, Tom Henighan,
  Shauna Kravec, Zac Hatfield-Dodds, Robert Lasenby, Dawn Drain, Carol Chen,
  Roger Gross, Sam McCandlish, Jared Kaplan, Dario Amodei, Martin Wattenberg,
  and Christopher Olah.
\newblock Toy models of superposition.
\newblock \emph{Transformer Circuits Thread}, 2022.
\newblock URL \url{https://transformer-circuits.pub/2022/toy_model/index.html}.

\bibitem[Engel et~al.(2001)Engel, Fries, and Singer]{engel2001dynamic}
Andreas~K. Engel, Pascal Fries, and Wolf Singer.
\newblock Dynamic predictions: Oscillations and synchrony in top-down
  processing.
\newblock \emph{Nature Reviews Neuroscience}, 2\penalty0 (10):\penalty0
  704--716, 2001.

\bibitem[Finn et~al.(2015)Finn, Shen, Scheinost, Rosenberg, Huang, Chun,
  Papademetris, and Constable]{finn2015fingerprinting}
Emily~S. Finn, Xilin Shen, Dustin Scheinost, Monica~D. Rosenberg, Jessica
  Huang, Marvin~M. Chun, Xenophon Papademetris, and R.~Todd Constable.
\newblock Functional connectome fingerprinting: Identifying individuals using
  patterns of brain connectivity.
\newblock \emph{Nature Neuroscience}, 18\penalty0 (11):\penalty0 1664--1671,
  2015.

\bibitem[Frank et~al.(2015)Frank, Otten, Galli, and Vigliocco]{frank2015erp}
Stefan~L. Frank, Leun~J. Otten, Giulia Galli, and Gabriella Vigliocco.
\newblock The {ERP} response to the amount of information conveyed by words in
  sentences.
\newblock \emph{Brain and Language}, 140:\penalty0 1--11, 2015.

\bibitem[Gao et~al.(2025)Gao, Ma, Chen, Li, Huang, Li, et~al.]{gao2025scaling}
Changjiang Gao, Zhengwu Ma, Jiajun Chen, Ping Li, Shujian Huang, Jixing Li,
  et~al.
\newblock Scaling, but not instruction tuning, increases large language models'
  alignment with language processing in the human brain.
\newblock \emph{Nature Computational Science}, 2025.
\newblock \doi{10.1038/s43588-025-00863-0}.

\bibitem[Goldstein et~al.(2022)Goldstein, Zada, Buchnik, Schain, Price, Aubrey,
  Nastase, Feder, Emanuel, Cohen, Jansen, Sompolinsky, and
  Hasson]{goldstein2022shared}
Ariel Goldstein, Zaid Zada, Eliav Buchnik, Mariano Schain, Amy Price, Bobbi
  Aubrey, Samuel~A. Nastase, Amir Feder, Dotan Emanuel, Alon Cohen, Aren
  Jansen, Haim Sompolinsky, and Uri Hasson.
\newblock Shared computational principles for language processing in humans and
  deep language models.
\newblock \emph{Nature Neuroscience}, 25\penalty0 (3):\penalty0 369--380, 2022.

\bibitem[Grandy et~al.(2013)Grandy, Werkle-Bergner, Chicherio, Schmiedek,
  L{\"o}vd{\'e}n, and Lindenberger]{grandy2013individual}
Thomas~H. Grandy, Markus Werkle-Bergner, Christian Chicherio, Florian
  Schmiedek, Michael L{\"o}vd{\'e}n, and Ulman Lindenberger.
\newblock Individual alpha peak frequency is related to latent factors of
  general cognitive abilities.
\newblock \emph{NeuroImage}, 79:\penalty0 10--18, 2013.

\bibitem[Grattafiori et~al.(2024)Grattafiori, Dubey, Jauhri, Pandey, Kadian,
  Al-Dahle, Letman, Mathur, Schelten, Vaughan, et~al.]{grattafiori2024llama}
Aaron Grattafiori, Abhimanyu Dubey, Abhinav Jauhri, Abhinav Pandey, Abhishek
  Kadian, Ahmad Al-Dahle, Aiesha Letman, Akhil Mathur, Alan Schelten, Alex
  Vaughan, et~al.
\newblock The {Llama} 3 herd of models.
\newblock \emph{arXiv preprint arXiv:2407.21783}, 2024.

\bibitem[Gratton et~al.(2018)Gratton, Laumann, Nielsen, Greene, Gordon,
  Gilmore, Nelson, Coalson, Snyder, Schlaggar, Petersen, and
  Dosenbach]{gratton2018functional}
Caterina Gratton, Timothy~O. Laumann, Ashley~N. Nielsen, Deanna~J. Greene,
  Evan~M. Gordon, Adrian~W. Gilmore, Steven~M. Nelson, Rebecca~S. Coalson,
  Abraham~Z. Snyder, Bradley~L. Schlaggar, Steven~E. Petersen, and Nico U.~F.
  Dosenbach.
\newblock Functional brain networks are dominated by stable group and
  individual factors, not cognitive or daily variation.
\newblock \emph{Neuron}, 98\penalty0 (2):\penalty0 439--452, 2018.

\bibitem[Hollenstein et~al.(2018)Hollenstein, Rotsztejn, Troendle, Pedroni,
  Zhang, and Langer]{hollenstein2018zuco}
Nora Hollenstein, Jonathan Rotsztejn, Marius Troendle, Andreas Pedroni,
  Ce~Zhang, and Nicolas Langer.
\newblock {ZuCo}, a simultaneous {EEG} and eye-tracking resource for natural
  sentence reading.
\newblock \emph{Scientific Data}, 5:\penalty0 180291, 2018.

\bibitem[Hollenstein et~al.(2019)Hollenstein, de~la Torre, Langer, and
  Zhang]{hollenstein2019cognival}
Nora Hollenstein, Antonio de~la Torre, Nicolas Langer, and Ce~Zhang.
\newblock {CogniVal}: A framework for cognitive word embedding evaluation.
\newblock In \emph{Proceedings of the 23rd Conference on Computational Natural
  Language Learning (CoNLL)}, 2019.

\bibitem[Hollenstein et~al.(2020)Hollenstein, Troendle, Zhang, and
  Langer]{hollenstein2020zuco}
Nora Hollenstein, Marius Troendle, Ce~Zhang, and Nicolas Langer.
\newblock {ZuCo} 2.0: A dataset of physiological recordings during natural
  reading and annotation.
\newblock In \emph{Proceedings of the 12th Language Resources and Evaluation
  Conference (LREC)}, 2020.

\bibitem[Hollenstein et~al.(2023)Hollenstein, Tr{\"o}ndle, Plomecka, Kiegeland,
  {\"O}zyurt, J{\"a}ger, and Langer]{hollenstein2023zuco}
Nora Hollenstein, Marius Tr{\"o}ndle, Martyna Plomecka, Samuel Kiegeland,
  Yilmazcan {\"O}zyurt, Lena~A J{\"a}ger, and Nicolas Langer.
\newblock The {ZuCo} benchmark on cross-subject reading task classification
  with {EEG} and eye-tracking data.
\newblock \emph{{Frontiers in Psychology}}, 13:\penalty0 1028824, 2023.

\bibitem[Jain \& Huth(2018)Jain and Huth]{jain2018incorporating}
Shailee Jain and Alexander Huth.
\newblock Incorporating context into language encoding models for fmri.
\newblock \emph{Advances in neural information processing systems}, 31, 2018.

\bibitem[Li et~al.(2023)Li, Patel, Vi{\'e}gas, Pfister, and
  Wattenberg]{li2023inference}
Kenneth Li, Oam Patel, Fernanda Vi{\'e}gas, Hanspeter Pfister, and Martin
  Wattenberg.
\newblock Inference-time intervention: Eliciting truthful answers from a
  language model.
\newblock \emph{Advances in Neural Information Processing Systems},
  36:\penalty0 41451--41530, 2023.

\bibitem[Marks \& Tegmark(2023)Marks and Tegmark]{marks2023geometry}
Samuel Marks and Max Tegmark.
\newblock The geometry of truth: Emergent linear structure in large language
  model representations of true/false datasets.
\newblock \emph{arXiv preprint arXiv:2310.06824}, 2023.

\bibitem[Michaelov et~al.(2024)Michaelov, Bardolph, Petten, Bergen, and
  Coulson]{michaelov2024strong}
James~A. Michaelov, Megan~D. Bardolph, Cyma K.~Van Petten, Benjamin~K. Bergen,
  and Seana Coulson.
\newblock Strong prediction: Language model surprisal explains multiple {N400}
  effects.
\newblock \emph{Neurobiology of Language}, 5\penalty0 (1):\penalty0 107--135,
  2024.

\bibitem[Nanda et~al.(2023)Nanda, Lee, and Wattenberg]{nanda2023emergent}
Neel Nanda, Andrew Lee, and Martin Wattenberg.
\newblock Emergent linear representations in world models of self-supervised
  sequence models.
\newblock In \emph{Proceedings of the 6th BlackboxNLP Workshop: Analyzing and
  Interpreting Neural Networks for NLP}, pp.\  16--30, 2023.

\bibitem[Panickssery et~al.(2023)Panickssery, Gabrieli, Schulz, Tong, Hubinger,
  and Turner]{panickssery2023steering}
Nina Panickssery, Nick Gabrieli, Julian Schulz, Meg Tong, Evan Hubinger, and
  Alexander~Matt Turner.
\newblock Steering {Llama} 2 via contrastive activation addition.
\newblock \emph{arXiv preprint arXiv:2312.06681}, 2023.

\bibitem[Park et~al.(2023)Park, Choe, and Veitch]{park2023linear}
Kiho Park, Yo~Joong Choe, and Victor Veitch.
\newblock The linear representation hypothesis and the geometry of large
  language models.
\newblock \emph{arXiv preprint arXiv:2311.03658}, 2023.

\bibitem[Pennington et~al.(2014)Pennington, Socher, and
  Manning]{pennington2014glove}
Jeffrey Pennington, Richard Socher, and Christopher~D. Manning.
\newblock {GloVe}: Global vectors for word representation.
\newblock In \emph{Proceedings of the 2014 Conference on Empirical Methods in
  Natural Language Processing (EMNLP)}, pp.\  1532--1543, 2014.

\bibitem[Qwen et~al.(2025)Qwen, Yang, Yang, Zhang, Hui, Zheng, Yu, Li, Liu,
  Huang, Wei, Lin, Yang, Tu, Zhang, Yang, Yang, Zhou, Lin, Dang, Lu, Bao, Yang,
  Yu, Li, Xue, Zhang, Zhu, Men, Lin, Li, Tang, Xia, Ren, Ren, Fan, Su, Zhang,
  Wan, Liu, Cui, Zhang, and Qiu]{qwen2025qwen25technicalreport}
Qwen, An~Yang, Baosong Yang, Beichen Zhang, Binyuan Hui, Bo~Zheng, Bowen Yu,
  Chengyuan Li, Dayiheng Liu, Fei Huang, Haoran Wei, Huan Lin, Jian Yang,
  Jianhong Tu, Jianwei Zhang, Jianxin Yang, Jiaxi Yang, Jingren Zhou, Junyang
  Lin, Kai Dang, Keming Lu, Keqin Bao, Kexin Yang, Le~Yu, Mei Li, Mingfeng Xue,
  Pei Zhang, Qin Zhu, Rui Men, Runji Lin, Tianhao Li, Tianyi Tang, Tingyu Xia,
  Xingzhang Ren, Xuancheng Ren, Yang Fan, Yang Su, Yichang Zhang, Yu~Wan,
  Yuqiong Liu, Zeyu Cui, Zhenru Zhang, and Zihan Qiu.
\newblock {Qwen2.5} technical report, 2025.
\newblock URL \url{https://arxiv.org/abs/2412.15115}.

\bibitem[Schrimpf et~al.(2021)Schrimpf, Blank, Tuckute, Kauf, Hosseini,
  Kanwisher, Tenenbaum, and Fedorenko]{schrimpf2021neural}
Martin Schrimpf, Idan~Asher Blank, Greta Tuckute, Carina Kauf, Eghbal~A.
  Hosseini, Nancy Kanwisher, Joshua~B. Tenenbaum, and Evelina Fedorenko.
\newblock The neural architecture of language: Integrative modeling converges
  on predictive processing.
\newblock \emph{Proceedings of the National Academy of Sciences (PNAS)},
  118\penalty0 (45):\penalty0 e2105646118, 2021.

\bibitem[Toneva \& Wehbe(2019)Toneva and Wehbe]{toneva2019interpreting}
Mariya Toneva and Leila Wehbe.
\newblock Interpreting and improving natural-language processing (in machines)
  with natural language-processing (in the brain).
\newblock In \emph{Advances in Neural Information Processing Systems
  (NeurIPS)}, 2019.

\bibitem[Tuckute et~al.(2024)Tuckute, Sathe, Srikant, Taliaferro, Wang,
  Schrimpf, Kay, and Fedorenko]{tuckute2024driving}
Greta Tuckute, Aalok Sathe, Shashank Srikant, Maya Taliaferro, Mingye Wang,
  Martin Schrimpf, Kendrick Kay, and Evelina Fedorenko.
\newblock Driving and suppressing the human language network using large
  language models.
\newblock \emph{Nature Human Behaviour}, 8:\penalty0 544--561, 2024.

\bibitem[Turner et~al.(2023)Turner, Thiergart, Leech, Udell, Vazquez, Mini, and
  MacDiarmid]{turner2023steering}
Alexander~Matt Turner, Lisa Thiergart, Gavin Leech, David Udell, Juan~J
  Vazquez, Ulisse Mini, and Monte MacDiarmid.
\newblock Steering language models with activation engineering.
\newblock \emph{arXiv preprint arXiv:2308.10248}, 2023.

\bibitem[Visconti~di Oleggio~Castello et~al.(2025)Visconti~di Oleggio~Castello,
  Dupr{\'e}~la Tour, and Gallant]{visconti2025individual}
Matteo Visconti~di Oleggio~Castello, Tom Dupr{\'e}~la Tour, and Jack~L.
  Gallant.
\newblock Individual differences shape conceptual representation in the brain.
\newblock \emph{bioRxiv}, 2025.
\newblock \doi{10.1101/2025.08.22.671848}.
\newblock URL
  \url{https://www.biorxiv.org/content/early/2025/08/22/2025.08.22.671848}.

\bibitem[Wolf et~al.(2020)Wolf, Debut, Sanh, Chaumond, Delangue, Moi, Cistac,
  Rault, Louf, Funtowicz, Davison, Shleifer, von Platen, Ma, Jernite, Plu, Xu,
  Scao, Gugger, Drame, Lhoest, and Rush]{wolf2020transformers}
Thomas Wolf, Lysandre Debut, Victor Sanh, Julien Chaumond, Clement Delangue,
  Anthony Moi, Pierric Cistac, Tim Rault, R{\'e}mi Louf, Morgan Funtowicz, Joe
  Davison, Sam Shleifer, Patrick von Platen, Clara Ma, Yacine Jernite, Julien
  Plu, Canwen Xu, Teven~Le Scao, Sylvain Gugger, Mariama Drame, Quentin Lhoest,
  and Alexander~M. Rush.
\newblock Transformers: State-of-the-art natural language processing.
\newblock In \emph{Proceedings of the 2020 Conference on Empirical Methods in
  Natural Language Processing: System Demonstrations (EMNLP)}, 2020.

\bibitem[Zhang et~al.(2024)Zhang, Rossi, Kveton, Shao, Yang, Zamani,
  Dernoncourt, Barrow, Yu, Kim, Zhang, Gu, Derr, Lipka, Sun,
  et~al.]{zhang2024personalization}
Zhehao Zhang, Ryan~A. Rossi, Branislav Kveton, Yijia Shao, Diyi Yang, Hamed
  Zamani, Franck Dernoncourt, Joe Barrow, Tong Yu, Sungchul Kim, Ruiyi Zhang,
  Jiuxiang Gu, Tyler Derr, Nedim Lipka, Tong Sun, et~al.
\newblock Personalization of large language models: A survey.
\newblock \emph{arXiv preprint arXiv:2411.00027}, 2024.

\bibitem[Zhao et~al.(2025)Zhao, Yan, Qiu, Ni, Zhang, Feng, Cheng, and
  Chua]{zhao2025steerx}
Xiaoyan Zhao, Ming Yan, Yilun Qiu, Haoting Ni, Yang Zhang, Fuli Feng, Hong
  Cheng, and Tat-Seng Chua.
\newblock {SteerX}: Disentangled steering for {LLM} personalization.
\newblock \emph{arXiv preprint arXiv:2510.22256}, 2025.

\bibitem[Zou et~al.(2023)Zou, Phan, Chen, Campbell, Guo, Ren, Pan, Yin,
  Mazeika, Dombrowski, et~al.]{zou2023representation}
Andy Zou, Long Phan, Sarah Chen, James Campbell, Phillip Guo, Richard Ren,
  Alexander Pan, Xuwang Yin, Mantas Mazeika, Ann-Kathrin Dombrowski, et~al.
\newblock Representation engineering: A top-down approach to {AI} transparency.
\newblock \emph{arXiv preprint arXiv:2310.01405}, 2023.

\end{thebibliography}
\end{document}